\documentclass{article}
\usepackage{spconf,epsfig}
\usepackage{amsmath,graphicx}
\usepackage{amsfonts,amssymb}
\usepackage{algorithm}
\usepackage[noend]{algpseudocode}
\usepackage[colorlinks=true, citecolor={blue}, bookmarks=false, breaklinks]{hyperref}
\usepackage{url}

\usepackage{breakurl}
\usepackage{multirow}
\usepackage{adjustbox}
\usepackage[numbers]{natbib}

\begin{document}\sloppy

\title{Use of a Capsule Network to Detect Fake Images and Videos}

\name{Huy H. Nguyen$^{\star}$, Junichi Yamagishi$^{\star\dagger\ddagger}$, and Isao Echizen$^{\star\dagger\S}$}
\address{$^{\star}$SOKENDAI (The Graduate University for Advanced Studies), Kanagawa, Japan\\
	$^{\dagger}$National Institute of Informatics, Tokyo, Japan\\
	$^{\S}$The University of Tokyo, Japan\\
	$^{\ddagger}$The University of Edinburgh, Edinburgh, UK\\
	\small{Email: \{nhhuy, jyamagis, iechizen\}@nii.ac.jp}}

\maketitle

\begin{abstract}
The revolution in computer hardware, especially in graphics processing units and tensor processing units, has enabled significant advances in computer graphics and artificial intelligence algorithms. In addition to their many beneficial applications in daily life and business, computer-generated/manipulated images and videos can be used for malicious purposes that violate security systems, privacy, and social trust. The deepfake phenomenon and its variations enable a normal user to use his or her personal computer to easily create fake videos of anybody from a short real online video. Several countermeasures have been introduced to deal with attacks using such videos. However, most of them are targeted at certain domains and are ineffective when applied to other domains or new attacks. In this paper, we introduce a capsule network that can detect various kinds of attacks, from presentation attacks using printed images and replayed videos to attacks using fake videos created using deep learning. It uses many fewer parameters than traditional convolutional neural networks with similar performance. Moreover, we explain, for the first time ever in the literature, the theory behind the application of capsule networks to the forensics problem through detailed analysis and visualization.
\end{abstract}


\section{Introduction}
Ever since the invention of photography, people have been interested in manipulating photographs, mainly to correct problems in the photos or to enhance them. However, we have gone far beyond these basic manipulations to adding unreal figures and inserting or removing objects. Digital photography has simplified the manipulation process, especially with the help of professional software like the iconic Adobe Photoshop application. The advent of personal computers further enabled people to become creators, creating everything from scratch. We call this computer graphics, as opposed to computer vision, which is aimed at making computers understand things from captured images and videos. The co-existence and co-development of computer graphics and computer vision, with the support of advanced hardware, have led to significant achievements. Moreover, the popularity of social networks has enabled people to create and share massive amounts of data, including personal information, news, and media, like images and videos. The consequence is that people with malicious purposes can easily make use of these advanced technologies and data to create fake images and videos and then publish them widely on social networks or use them to bypass facial authentication.

\begin{figure}[t!]
\centering
\includegraphics[width=85mm]{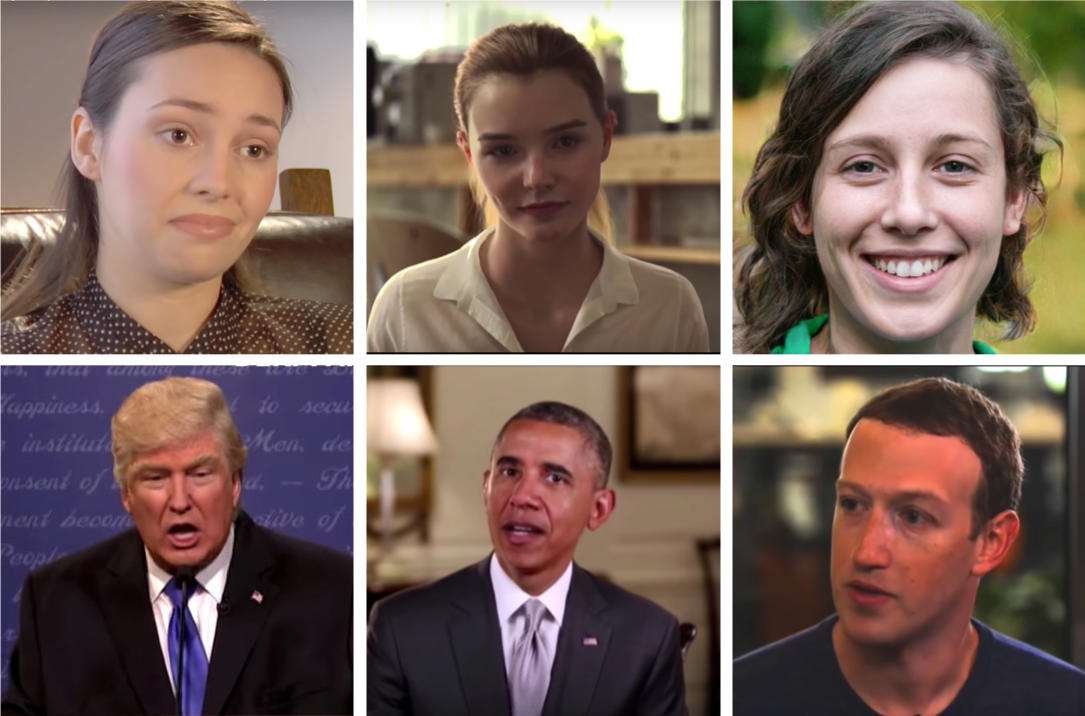}
\caption{Example fake images. The first two images in the top row were fully computer-generated (from the Digital Emily Project~\cite{alexander2010digital} and from Dexter Studios~\cite{dexterstudio}); the last one was generated using StyleGAN~\cite{karras2019style}. In the bottom row, from left to right, are images manipulated using deepfake~\cite{deepfake}, Face2Face~\cite{thies2016face2face}, and Neural Textures~\cite{thies2019deferred} methods, respectively.}
\label{fig:example}
\end{figure}

The requirements for manipulating or synthesizing videos were dramatically simplified when it became possible to create forged videos from only a short video of the target person~\cite{thies2016face2face, kim2018deep} and then from a single ID photo~\cite{averbuch2017bringing} following the acting of an actor. Suwajanakorn et al.'s mapping method~\cite{suwajanakorn2017synthesizing} enhanced the ability of manipulators to learn the mapping between speech and lip motion. Since state-of-the-art speech synthesis methods can produce natural sounding speech, this mapping method enabled the creation of a fully synthesized audio-video image of any person. Deepfakes~\cite{deepfake} exemplify this threat -- any user with a personal computer and an appropriate tool can create videos impersonating any celebrity. Since deepfakes have become easy to create, a large number of high-quality fake pornography videos have been produced. Deepfake comedy and deepfake videos have been posted on YouTube with the challenge being to spot them. Several examples of computer-generated/manipulated images are shown in Fig.~\ref{fig:example}.

Several countermeasures have been developed for detecting spoofing attacks using fake images and videos. Before the deepfake phenomenon~\cite{deepfake}, when computer-generated images and videos had not yet achieved the required realistic quality to become a threat, presentation attacks~\cite{marcel2019handbook} were the main concern, and the detectors used hand-crafted features~\cite{chingovska2012effectiveness, de2013can, kim2015face}. The growing use of convolutional neural networks (CNNs) has changed the game drastically for both defenders and attackers. Automatic feature extraction has dramatically improved detection performance~\cite{ito2017recent, george2019deep} while deep generative methods like generative adversarial networks (GANs) enable images~\cite{karras2019style} and videos~\cite{tripathy2019icface, yuval2019fsgan} to be produced that are almost humanly impossible to detect as fake.
The attention of the forensics community has thus shifted to these new kinds of attacks~\cite{li2018ictu, korshunovvulnerability, agarwal2019protecting, sabir2019recurrent}. Several approaches are image-based~\cite{afchar2018mesonet, rossler2018faceforensics} while others work only on videos frames~\cite{li2018ictu, agarwal2019protecting, sabir2019recurrent} or on video frames and voice information~\cite{korshunov2018speaker}. Although some video-based approaches perform better than image-based ones, they are only applicable to particular kinds of attacks. For example, many of them~\cite{li2018ictu, agarwal2019protecting} may fail if the quality of the eye area is sufficiently good or the synchronization between the video and audio parts is sufficiently natural~\cite{korshunovvulnerability}. Among the image-based approaches, some are general-purpose detectors, for instance, Fridrich and Kodovsky's one~\cite{fridrich2012rich} (applicable to both steganalysis and facial reenactment video detection) and Rahmouni et al.'s one~\cite{rahmouni2017distinguishing} (applicable initially to computer-generated images and later to computer-manipulated images). However, their performance on new tasks is limited compared with that of task-specific ones~\cite{rossler2018faceforensics, rossler2019faceforensics++}.

This journal paper is an extension of our conference paper~\cite{nguyen2019capsule} in which we pioneered the use of capsule networks~\cite{hinton2011transforming, sabour2017dynamic} for digital media forensics problems. We aim to create a lightweight and general-purpose detector that can be used for any kind of attack and have reasonable performance compared with that of task-specific detectors. While most state-of-the-art detectors use traditional CNNs with a large number of parameters, ours uses a new type of CNN that has impressive performance on computer vision tasks. The network architecture is relatively new and has seen little application in other domains, and a detailed analysis of it has been lacking. To fill this gap, we explain the novelty of our proposed capsule network through detailed analysis and visualization of several kinds of attacks. We also describe how we enhanced its performance by making several modifications and introducing two regularizations.

\section{Related Work}
In this section, we first introduce several state-of-the-art face manipulation techniques that can be used to manipulate faces. We then mention several major research efforts focused on forgery detection during the eight years preceding our research, the time period when CNNs blossomed and came to dominate traditional methods. As best we can, we group them into presentation attack detection and computer-generated image/video detection on the basis of the features they use and their intended targets. Despite this categorization, some approaches are two-fold or can be successfully applied outside their original scopes. Finally, we provide basic information about capsule networks, the dynamic routing algorithm that enables them to be efficiently implemented, and their original application to computer vision.

\subsection{Face Manipulation}
Although face manipulation is not new, recent achievements demonstrate that computer-manipulated faces can reach a photo-realistic level at which it is almost impossible for them to be humanly detected as fake. Dale et al.~\cite{dale2011video} presented a 3D multilinear model for replacing facial movements in video. Garrido et al.~\cite{garrido2015vdub} modified the lip movements of an actor in a target video so that they matched a different audio track. Thies et al.~\cite{thies2016face2face} demonstrated that expression transfer for facial reenactment can be performed in real time and subsequently developed the FaceVR algorithm~\cite{thies2016facevr}, which handles eye-tracking and reenactment in virtual reality. Kim et al.~\cite{kim2018deep} demonstrated the transfer of a head pose along with facial movements from an actor to another person. Similarly, Tripathy et al.~\cite{tripathy2019icface} devised a lightweight face reenactment method using GANs. Nirkin et al.~\cite{yuval2019fsgan} presented a face swapping method that does not require training on new faces, unlike deepfake methods~\cite{deepfake}. Thies et al. combined the traditional graphics pipeline with learnable components to deal with imperfect 3D contents~\cite{thies2019deferred}.

Not only visual part, Suwajanakorn et al.~\cite{suwajanakorn2017synthesizing} presented a method for learning the mapping between speech and lip movements in which speech can also be synthesized, enabling creation of a full-function spoof video. Fried et al.~\cite{fried2019text} demonstrated that speech can be easily modified in any video in accordance with the intention of the manipulator while maintaining a seamless audio-visual flow. Averbuch-Elor et al.~\cite{averbuch2017bringing} addressed a different problem -- converting still portraits into motion pictures expressing various emotions. This work greatly simplified the requirements for attackers: simply acquire a picture of the victim (usually a profile picture on a social network or an ID photo). Zakharov et al.~\cite{zakharov2019few} followed up by improving the quality of videos generated using only a few input images. Vougioukas et al.~\cite{vougioukas2019end} raised the bar by introducing a method for animating a facial image from an audio track containing speech. 

Besides these academic efforts, deep-learning-based face swapping tools (like deepfake's source code\footnote{https://github.com/deepfakes/faceswap}) have become widespread on the Internet, enabling normal users to create pornographic videos with celebrity images or to impersonate them. For users who are not familiar with programming and machine learning, there is a mobile app called Xpression\footnote{https://xpression.jp/} that provides a deepfake function. An even more controversial application recently appeared -- DeepNude~\cite{deepnude}, which generates realistic nude images of a person from a picture of him or her wearing clothes. It was shut down hours after it was released due to negative reaction from the community. 

\subsection{Presentation Attack Detection}
A presentation attack against a biometric capture system is an attack with the goal of interfering with the system's operation. Presentation attack detection (PAD) methods have been developed to automatically detect this kind of attack. Local binary patterns and their variances were the most effective PAD features in the pre-deep learning era and are used in several methods~\cite{chingovska2012effectiveness, de2013can, kim2015face}. Following the success of methods based on CNNs in the ImageNet Large Scale Visual Recognition Challenge (ILSVRC)~\cite{ILSVRC15}, several methods have been developed that leverage the pre-trained CNNs in this Challenge's large database as their feature extractors~\cite{yang2014learn, ito2017recent}. Other methods have been developed that use the available CNN architectures with customized components and were trained on spoofing databases~\cite{menotti2015deep, raghavendra2017transferable, george2019deep, mehtacrafting, muhammad2019face}. Besides that, Alotaibi and Mahmood~\cite{alotaibi2017deep} applied nonlinear diffusion based on an additive operator splitting scheme to their CNN.

There is growing interest in generalizing detectors to enable them to handle unseen attacks~\cite{costa2019generalized}. This is a difficult but important effort since the number of attack techniques and their variances have been increasing rapidly. The major directions for countermeasure are using adversarial training and domain-adaptation jointly or independently~\cite{jaiswal2019ropad, nikisins2019domain, wangimproving}. Other directions are semi-supervised learning~\cite{cozzolino2018forensictransfer} and semi-supervised learning combined with multi-task learning~\cite{nguyen2019multi}. On the other hand, Fatemifar et al.~\cite{fatemifar2019combining} applied a fusion-based multiple one-class classifier approach to anomaly detection.

\subsection{Computer-Generated/Manipulated Image/Video Detection}
Although computer-manipulation of images and videos is nothing new, the introduction of deep learning significantly improved the ability of this kind of attack and thus attracted great attention from the forensics community. This resulted in the creation of standardized databases for benchmarking, like the FaceForensics~\cite{rossler2018faceforensics}, FaceForensics++~\cite{rossler2019faceforensics++}, and DeepFakeTIMIT databases~\cite{korshunovvulnerability}. These databases cover several well-known attacks, including Face2Face~\cite{thies2016face2face}, FaceSwap~\cite{rossler2019faceforensics++}, and deepfake~\cite{deepfake}. Rahmouni et al. had previously created a database for detecting fully computer-generated images~\cite{rahmouni2017distinguishing} while Afchar et al.~\cite{afchar2018mesonet} created a deepfake database in their pioneering deepfake detection work.

The handcrafted steganalysis-based method developed by Fridrich and Kodovsky~\cite{fridrich2012rich} was used in early efforts to detect computer-manipulated images and videos. This approach was later implemented in a CNN by Cozzolino et al.~\cite{cozzolino2017recasting}. Subsequently, several methods based on CNNs which have been used in the ILSVRC, like the one developed by Rossler et al.~\cite{rossler2018faceforensics}. Other methods used networks proposed by their authors~\cite{bayar2016deep, rahmouni2017distinguishing, quan2018distinguishing, afchar2018mesonet, li2018ictu,korshunov2018speaker,sabir2019recurrent} while others are based on a hybrid approach~\cite{zhou2017two, nguyen2018modular, wang2019detecting, nguyen2019capsule}. Beside deep-learning and non-deep-learning categorization, these methods could be divided into image-based classifiers~\cite{bayar2016deep, zhou2017two, rahmouni2017distinguishing, quan2018distinguishing, afchar2018mesonet, rossler2018faceforensics, nguyen2018modular, wang2019detecting, nguyen2019capsule} and video-based classifiers~\cite{li2018ictu, korshunov2018speaker, agarwal2019protecting, sabir2019recurrent}. For detecting images generated by GANs, Marra et al.~\cite{marra2018detection} performed benchmark testing on several CNNs and proposed a statistical model for detection~\cite{marra2019gans}.

In addition to binary classification between real and modified/generated images or videos, locating manipulated regions in images is also a major branch in digital media forensics. This research focuses on detecting removal, copy-move, and splicing attacks. Besides forensic-oriented approaches~\cite{bappy2017exploiting, zhou2018learning, bappy2019hybrid, nguyen2019multi}, semantic segmentation approaches~\cite{long2015fully, badrinarayanan2017segnet} and binary classification approaches are applicable~\cite{rahmouni2017distinguishing, nguyen2018modular, rossler2018faceforensics, rossler2019faceforensics++}. In the case of binary classifiers, a sliding window (with or without overlapping) is used to locate manipulated regions.

\subsection{Capsule Networks}
``Capsule network'' is not the new term as it was first introduced in 2011 by Hinton et al.~\cite{hinton2011transforming}. They argued that CNNs have limited applicability to the ``inverse graphics'' problem and introduced a more robust architecture comprising several ``capsules.'' However, they initially faced the same problem faced by CNNs -- the limited performance of hardware and the lack of effective algorithms, which prevented practical application of capsule networks. CNNs thus remained dominant in this research field.

These problems were overcome when the dynamic routing algorithm~\cite{sabour2017dynamic} and its variance -- the expectation-maximization routing algorithm~\cite{hinton2018matrix} -- were introduced in 2017 and 2018, respectively. These breakthroughs enabled capsule networks to achieve better performance and outperform CNNs on object classification tasks~\cite{sabour2017dynamic, xi2017capsule, hinton2018matrix, xiang2018ms, bahadori2018spectral}. The agreements between low- and high-level capsules that encode the hierarchical relationships between objects and their parts with pose information enables a capsule network to preserve more information than a CNN while using only a fraction of the data used by a CNN.

In another domain, Iesmantas and Alzbutas applied a capsule network based on binary classification to breast cancer detection~\cite{iesmantas2018convolutional}. Jaiswal et al. reported a capsule-based GAN~\cite{jaiswal2018capsulegan}. Yang et al. applied a capsule network to the text domain~\cite{yang2018investigating}. Nguyen et al. were pioneers in applying capsule networks to the digital media forensics problem~\cite{nguyen2019capsule}. These efforts demonstrated the effectiveness of capsule networks in multiple domains, which motivated us to continue developing a method for detecting modified/generated images or videos that we call ``Capsule-Forensics'' and then to perform a deep analysis on its behaviors by visualizing its intermediate activations on both real and fake inputs to explain the theory behind it.

\section{Capsule-Forensics}
\subsection{Overview}
\begin{figure}[th!]
\centering
\includegraphics[width=86mm]{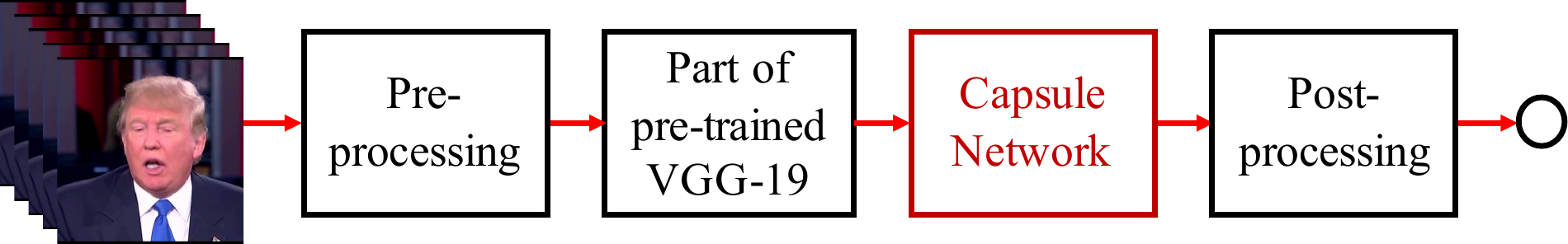}
\caption{Capsule-Forensics pipeline.}
\label{fig:overview}
\end{figure}

The pipeline of our proposed Capsule-Forensics method is illustrated in Fig.~\ref{fig:overview}. The pre-processing task depends on the input. If the input is video, the first step is to separate the frames. If the task is to detect fully computer-generated frames (or images), each frame (or image) is divided into patches. If the task is to detect a fake face or faces, a face detection algorithm is used to crop the facial area(s). There is no strict requirement about the size of the output image. In general, the larger the input, the better the result, at the cost of more computational power. The commonly used image sizes in practice are $100 \times 100$, $128 \times 128$, $256 \times 256$, and $299 \times 299$~\cite{rahmouni2017distinguishing, nguyen2019capsule, nguyen2018modular, rossler2018faceforensics, rossler2019faceforensics++}. We used an image size of $300 \times 300$ as it is an even number (making it easy to perform cropping and scaling) and large enough to provide sufficient information for detecting fake content.

\begin{figure*}[th!]
\centering
\includegraphics[width=165mm]{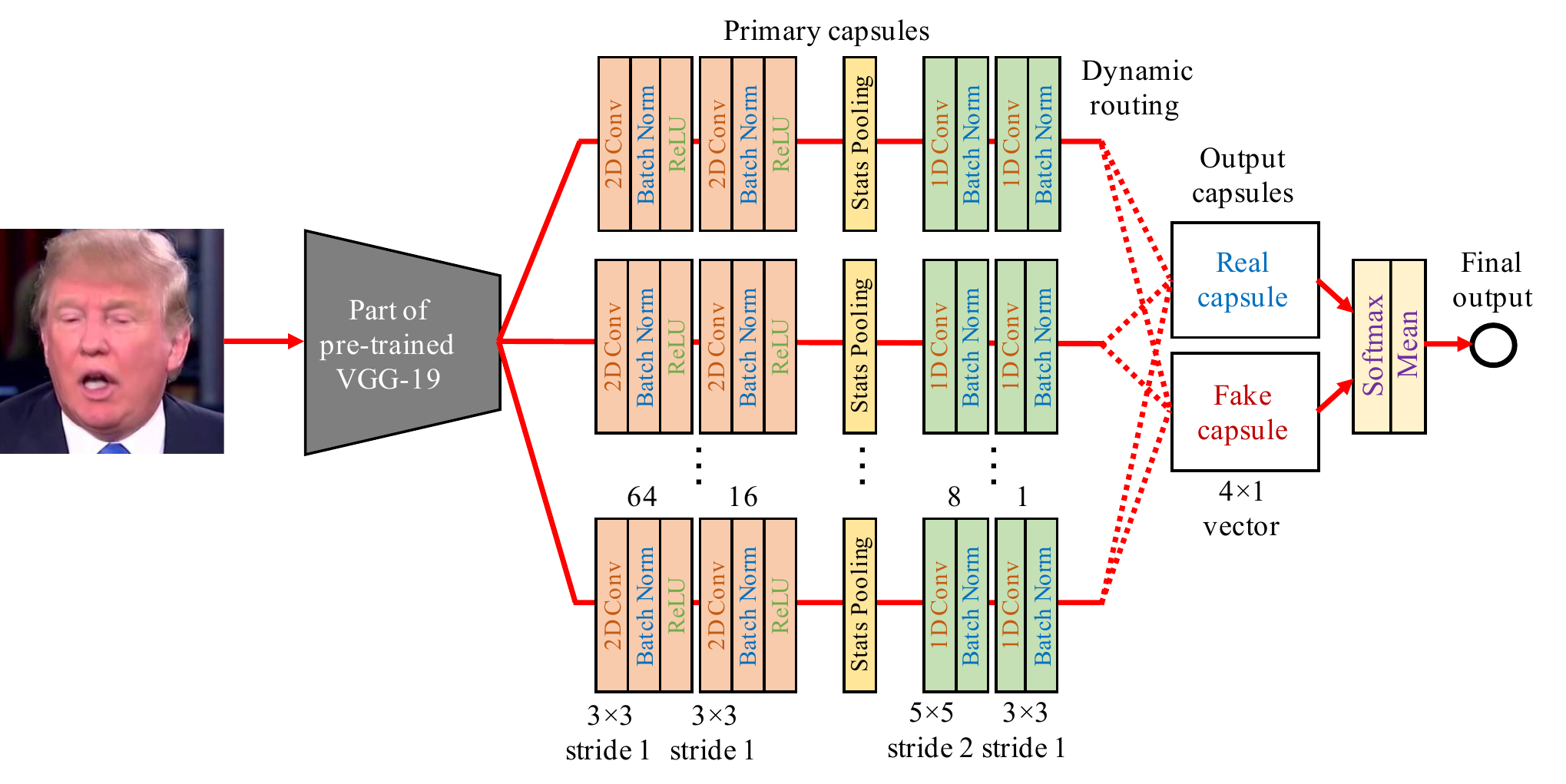}
\caption{Capsule-Forensics architecture.}
\label{fig:capsnet}
\end{figure*}

The pre-processed image then passes through a part of the VGG-19 network (as proposed by the Visual Geometry Group)~\cite{simonyan2014very} pre-trained on the ILSVRC database~\cite{ILSVRC15} before entering the capsule network. The VGG-19 network is used from the first layer to the third max pooling layer, which is not too deep to obtain biases from the object detection task (the original purpose of this pre-trained network). This VGG-19 part is equivalent to the CNN part before the primary capsules in the design of the original capsule network~\cite{sabour2017dynamic}. Using a pre-trained CNN as a feature extractor rather than training it from scratch provides the benefit of using it as a regularizer to guide the training and to reduce overfitting as well as that of transfer learning. The detailed architecture is discussed in the next section.

The final part is the post-processing unit, which works in accordance with the pre-processing one. If the task is to detect fully computer-generated images, the scores of the extracted patches are averaged. If the input is video, the scores of all frames are averaged. This average score is the final output.

\subsection{Detailed Architecture}
The capsule network includes several primary capsules and two output capsules (``real'' and ``fake''), as illustrated in Fig.~\ref{fig:capsnet}. There is no constraint on the number of primary capsules. Experiments demonstrated that a reasonably large number of primary capsules may improve network performance, but at the cost of more computation power. Three capsules are typically used for light networks (which require less memory and computation), and ten capsules are typically used for full ones (which require more memory and computation but provide better performance). While it is not necessary to use the same architecture for the primary capsules, we used the same design for all primary capsules to simplify the discussion.

Each primary capsule is divided into three parts: a 2D convolutional part, a statistical pooling layer, and a 1D convolutional part. The statistical pooling layer has been proven to be effective in the forensics task~\cite{rahmouni2017distinguishing, nguyen2018modular}. Moreover, it helps make the network independent of the input image size. This means that one Capsule-Forensics architecture can be applied to different problems with different input sizes without having to redesign the network. The mean and variance of each filter are calculated in the statistical pooling layer.
\begin{itemize}
\item Mean: $$\mu_k = \frac{1}{H\times W}\sum_{i=1}^{H}\sum_{j=1}^{W}I_{kij}$$.
\item Variance: $$\sigma_k^2 = \frac{1}{H\times W-1}\sum_{i=1}^{H}\sum_{j=1}^{W}(I_{kij}-\mu_k)^2$$,
\end{itemize}
where $k$ represents the layer index, $H$ and $W$ are respectively the height and width of the filter, and $I$ is a two-dimensional filter array.

The output of the statistical layer is suitable for 1D convolution. After going through the following 1D convolutional part, it is sent through dynamic routing to the output capsules. The final result is calculated on the basis of the activation of the output capsules. The algorithm is discussed in detail in the next section. For binary classification, there are two output capsules, as shown in Fig.~\ref{fig:capsnet}. Multi-class classification could be performed by adding more output capsules, as described in section~\ref{sec:faceforensics}.

\subsection{Dynamic Routing Algorithm}
The dynamic routing algorithm is used to calculate agreement between the features extracted by the primary capsules. Agreement is dynamically calculated at run-time and the results are routed to the appropriate output capsule (real or fake one for binary classification). The output probabilities are determined on the basis of the activations of the output capsules. This dynamic routing algorithm differs from the classical fusion one in that it combines classification outputs from different classifiers.

Let us call the output vector of each primary capsule $\textbf{u}^{(i)}$ and the real and fake vector capsules $\textbf{v}^{(1)}$ and $\textbf{v}^{(2)}$, respectively. $\mathrm{W}^{(i,j)}$ is the matrix used to route $\textbf{u}^{(i)}$ to $\textbf{v}^{(j)}$, and $r$ is the number of iterations. The dynamic routing algorithm is shown in Algorithm~\ref{alg:routing}.

\begin{algorithm}
	\caption{Dynamic routing between capsules.}
	\label{alg:routing}
	\begin{algorithmic}
		\Procedure{Routing}{$\textbf{u}^{(i)}, \mathrm{W}^{(i,j)}, r$}
		\State $\hat{\mathrm{W}}^{(i,j)}\gets \mathrm{W}^{(i,j)} + \text{rand}(\text{size}(\mathrm{W}^{(i,j)}))$
		\State $\hat{\textbf{u}}^{(i)}\gets \hat{\mathrm{W}}^{(i,j)} \text{squash}(\textbf{u}^{(i)})$
		\State $\hat{\textbf{u}}^{(i)}\gets \text{dropout}(\hat{\textbf{u}}^{(i)})$
		\For {all input capsule $i$ and all output capsules $j$}
		\State $b_{i,j}\gets 0$
		\EndFor
		\For {$r$ iterations}
		\State \textbf{for} all input capsules $i$ \textbf{do} $\textbf{c}_{i}\gets \text{softmax}(\textbf{b}_{i})$
		\State \textbf{for} all output capsules $j$ \textbf{do} $\textbf{s}_{j}\gets \sum_{i} c_{i,j}\hat{\textbf{u}}^{(i)}$
		\State \textbf{for} all output capsules $j$ \textbf{do} $\textbf{v}^{(j)}\gets \text{squash}(\textbf{s}_{j})$
		\For {all input capsules $i$ and output capsules $j$}
		\State $b_{(i,j)}\gets b_{i,j} + \hat{\textbf{u}}^{(i)\intercal} \textbf{v}^{(j)}$
		\EndFor
		\EndFor
		\State \textbf{return} $\textbf{v}^{(j)}$
		\EndProcedure
	\end{algorithmic}
\end{algorithm}

We slightly improved the algorithm of Sabour et al.~\cite{sabour2017dynamic} by introducing two regularizations: adding random noise to the routing matrix and adding a dropout operation. They are used \textbf{only during training} to reduce overfitting. Their effectiveness is discussed in the Evaluation section. Furthermore, a squash function (equation~\ref{eq:squash}) is applied to $\textbf{u}^{(i)}$ before routing to normalize it, which helps stabilize the training process. The squash function is used to scale the vector magnitude to unit length.

\begin{equation}
	\label{eq:squash}
	squash(\textbf{u}) = \frac{\|\textbf{u}\|_2^2}{1 + \|\textbf{u}\|_2^2}\frac{\textbf{u}}{\|\textbf{u}\|_2}
\end{equation}

In practice, to stabilize the training process, the random noise should be sampled from a normal distribution ($\mathcal{N}(0, 0.01)$), the dropout ratio should not be greater than 0.05 (we used 0.05 in all experiments), and two iterations ($r = 2$) should be used in the dynamic routing algorithm. The two regularizations are used along with random weight initialization to increase the level of randomness, which helps the primary capsules to learn with different parameters.

To calculate predicted label $\hat{y}$, we apply the softmax function to each dimension of the output capsule vectors to achieve stronger polarization rather than simply using the length of the output capsules~\cite{sabour2017dynamic}. The final results are the means of all softmax outputs:

\begin{equation}
	\label{eq:predict}
	\hat{\textbf{y}} = \frac{1}{m} \sum_i \text{softmax} \left({\begin{bmatrix} \textbf{v}^{(1)\intercal} \\ \textbf{v}^{(2)\intercal} \end{bmatrix}}_{:,i}\right).
\end{equation}

Since there is no reconstruction in Capsule-Forensics, we simply use the cross-entropy loss function (equation~\ref{eq:loss}) and the Adam optimizer~\cite{kingma2014adam} to optimize the network:

\begin{equation}
	\label{eq:loss}
	L = -\left(y\log(\hat{y}) + (1 - y)\log(1 - \hat{y}) \right),
\end{equation}
where $y$ is the ground truth label, $\hat{y}$ is the predicted label, and $m$ is the dimensional of the output capsule $\textbf{v}_j$.

\subsection{How Capsule-Forensics Works}
To illustrate how Capsule-Forensics works, we used a Capsule-Forensics network with three primary capsules trained on the FaceForensics++ database~\cite{rossler2019faceforensics++}. We applied both regularizations (using random noise and dropout during training) and used images cropped to $300 \times 300$. For visualization, we applied and modified an open-source tool~\cite{uozbulak_pytorch_vis_2019} implementing the guided back-propagation algorithm~\cite{selvaraju2017grad}. To visualize each primary capsule in this way, we chose the latent features extracted before the statistical pooling layers since they still had the 2D structure.

\begin{figure*}[th!]
\centering
\includegraphics[width=108mm]{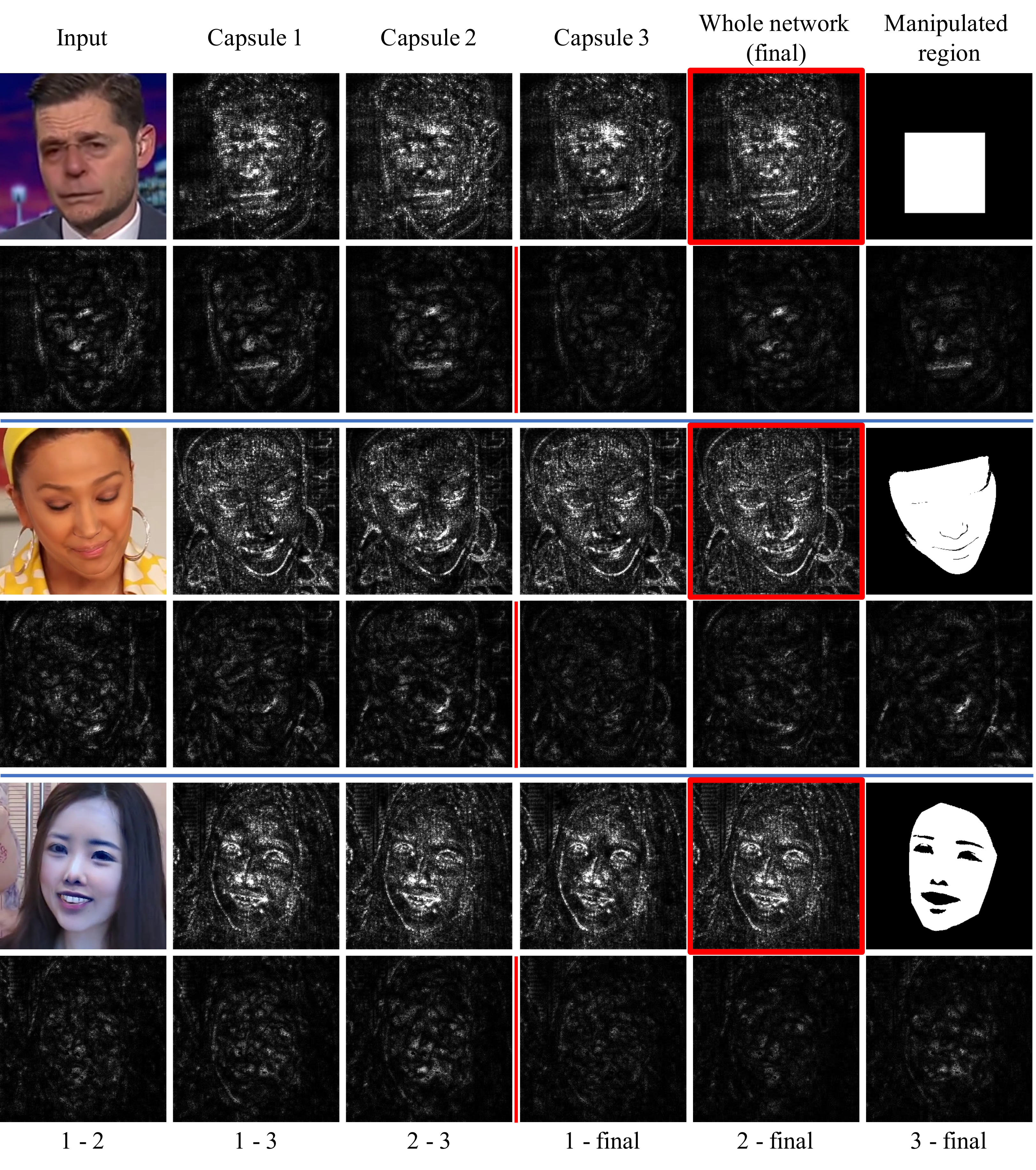}
\caption[caption]{Activation of the three capsules and the whole Capsule-Forensics network (columns 2, 3, 4, and 5, respectively) on images created using deepfake~\cite{deepfake} (row 1), Face2Face~\cite{thies2016face2face} (row 3), and FaceSwap~\cite{rossler2019faceforensics++} (row 5) methods. Column 6 shows the manipulated regions corresponding to the manipulated images in column 1.\\\hspace{\textwidth}The first three columns of rows 2, 4, and 6 show the differences between the activations of capsules 1 and 2, 1 and 3, and 2 and 3 on the corresponding row above, respectively. The three last columns in order show the differences between the activations of capsules 1, 2, and 3 and the activation of the whole network.}
\label{fig:viz}
\end{figure*}

\begin{figure*}[th!]
\centering
\includegraphics[width=76mm]{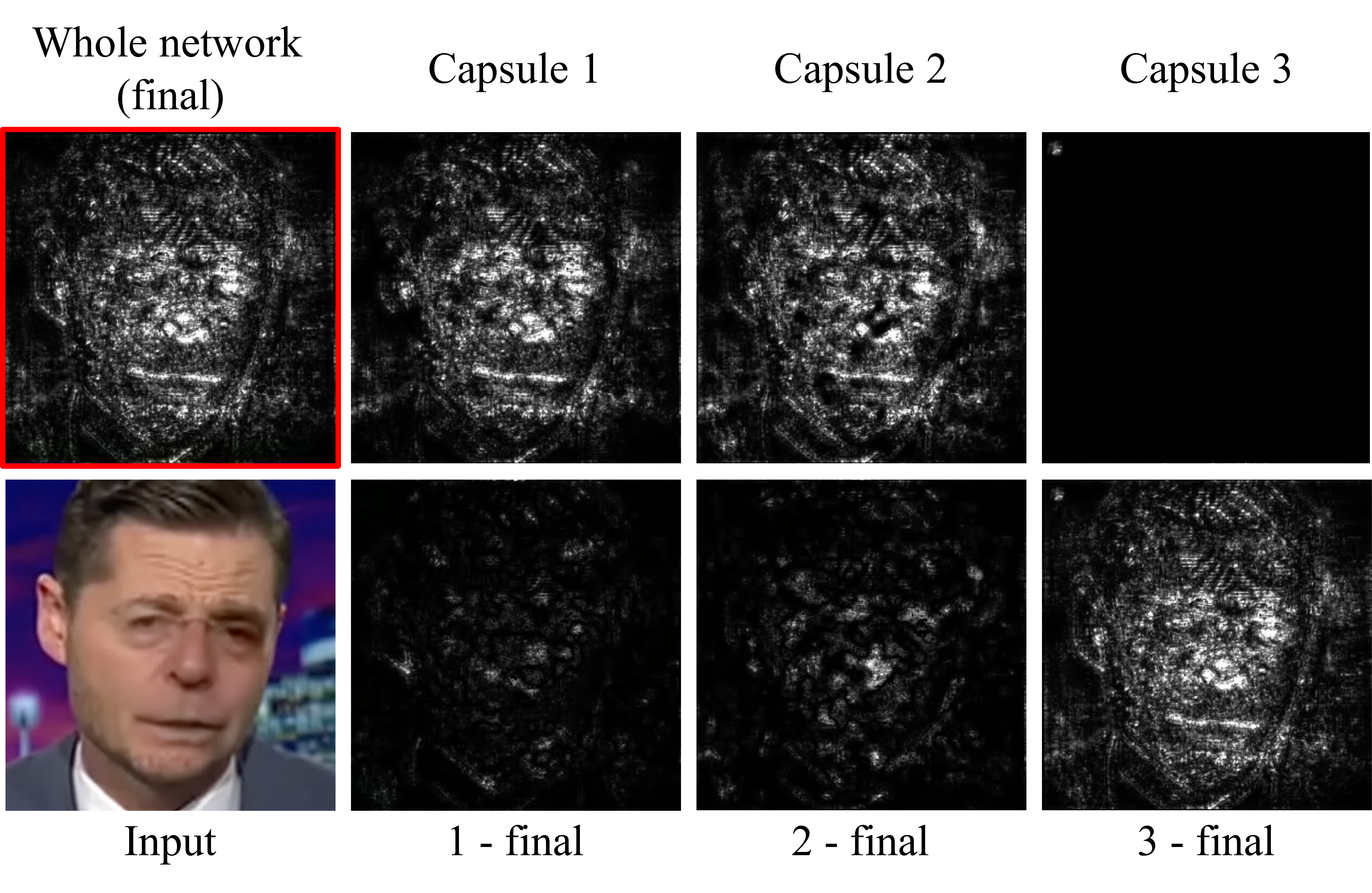}
\caption{Example case in which one capsule did not work correctly. The first row shows the activation of the whole network and of the three capsules. The second row from left to right shows the input image and the differences between the activation of each capsule and of the whole network. Capsule 3 failed to detect the manipulated image, but thanks to the other two capsules, the final result was still correct.}
\label{fig:viz_global_fail}
\end{figure*}

\textbf{The first question} we address is about capsule learning: What did each capsule learn and were the learned features the same given that capsules had the same architecture? Before training a neural net or a capsule network, weight initialization needs to be applied (in our case, we used a normal distribution for weight ninitialization). Therefore, their starting points differed. During the learning process, these initial differences forced each capsule to focus on features that may be near but not identical to the others. The activation of each capsule and of the whole network are illustrated in Fig.~\ref{fig:viz}. The differences in activation among capsules and between each capsule and the whole network are also shown. The regions of interest mainly include the eyes, nose, mouth region, and facial contours. Several capsules missed some of these regions, and several failed to detect the manipulated input (e.g., the 3\textsuperscript{rd} capsule in Fig~\ref{fig:viz_global_fail}). However, thanks to the agreements between the capsules driven by the dynamic routing algorithm, the final results mostly focused on the important regions detected by all capsules. If the Capsule-Forensics network was replaced by a CNN using only the third primary capsule, the CNN would also fail to detect the manipulated input.

Since detecting manipulated images is the forensics problem addressed here, the behavior of the Capsule-Forensics network is differs from that in the explanation of the original capsule network for the inverse graphics problem, in which the focus is on the spatial hierarchies between simple and complex objects~\cite{hinton2011transforming,sabour2017dynamic,hinton2018matrix}. In the forensics problem, abnormal appearances are the key features, so each primary capsule tries its best to capture them and communicate its findings to the other capsules. This behavior is similar to that of jurors during a trial, and the judgment is the final detection result.

\begin{figure*}[th!]
\centering
\includegraphics[width=100mm]{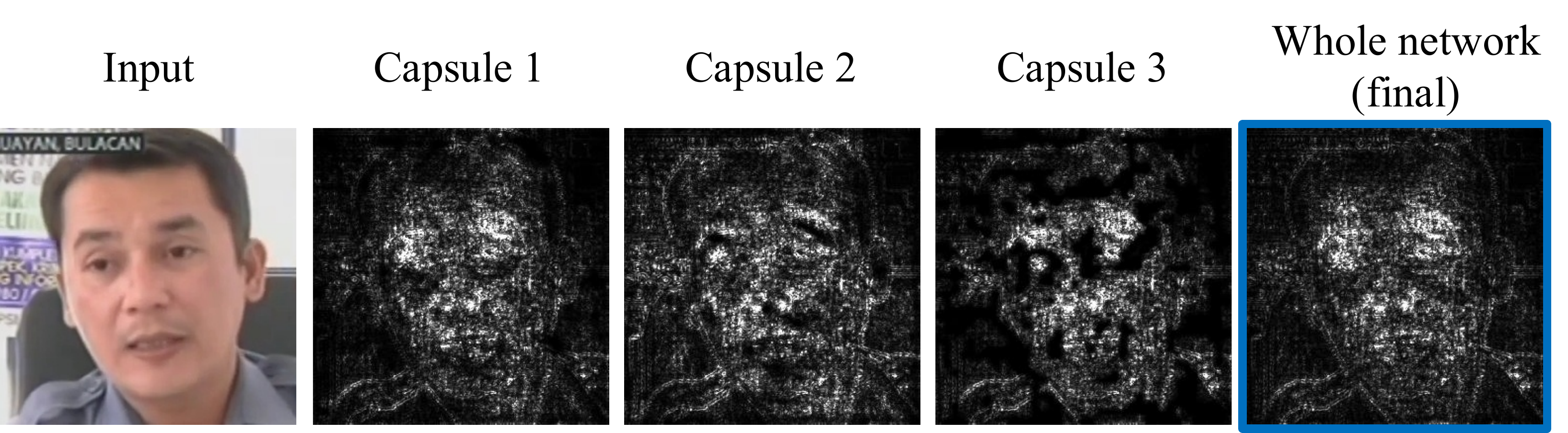}
\caption{Activation of three primary capsules and of whole network for real input.}
\label{fig:viz_real}
\end{figure*}

\begin{figure*}[th!]
\centering
\includegraphics[width=140mm]{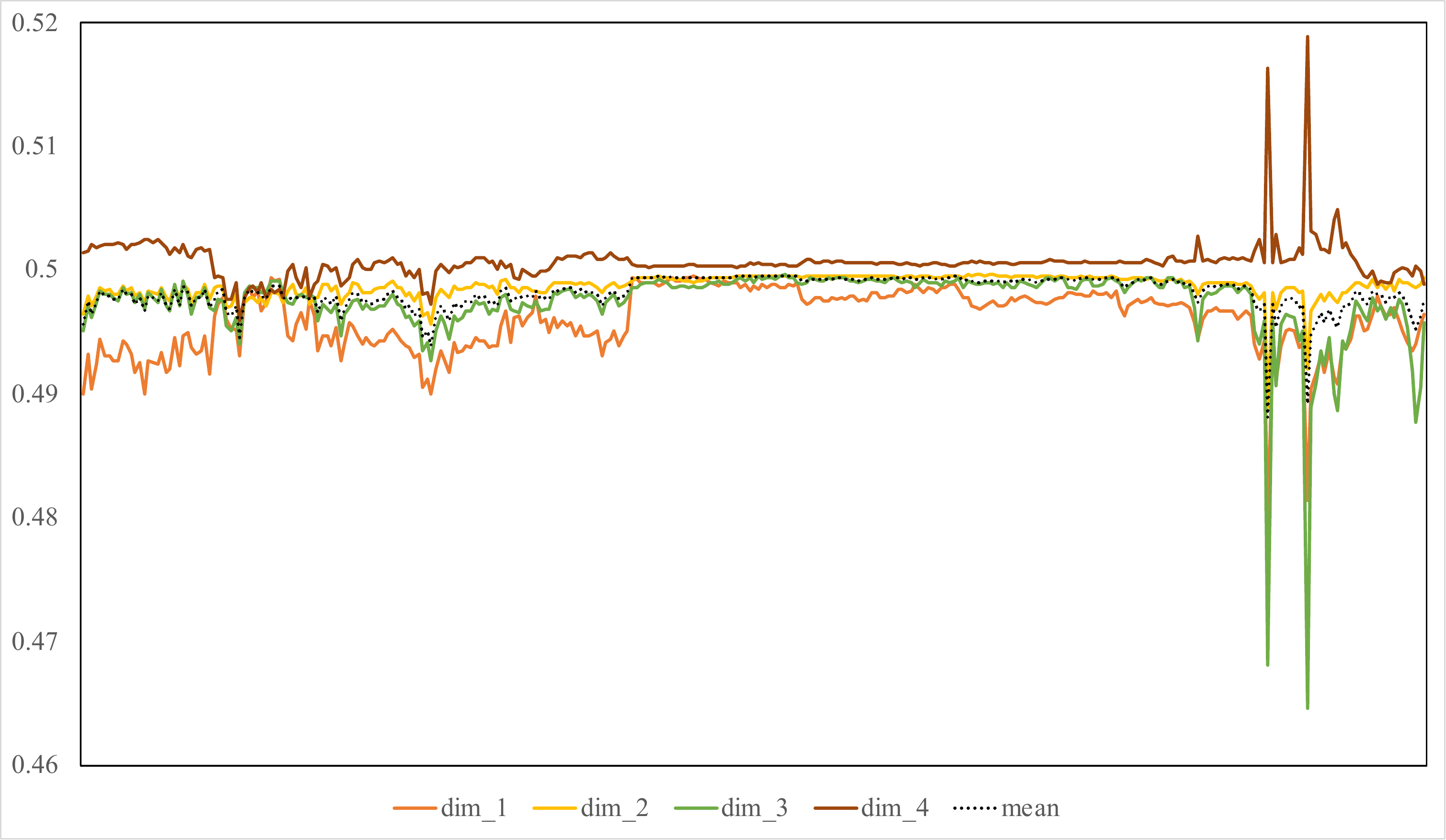}
\caption{Agreement between three primary capsules and fake output capsule for fake input.}
\label{fig:agreement_real}
\end{figure*}

\begin{figure*}[th!]
\centering
\includegraphics[width=140mm]{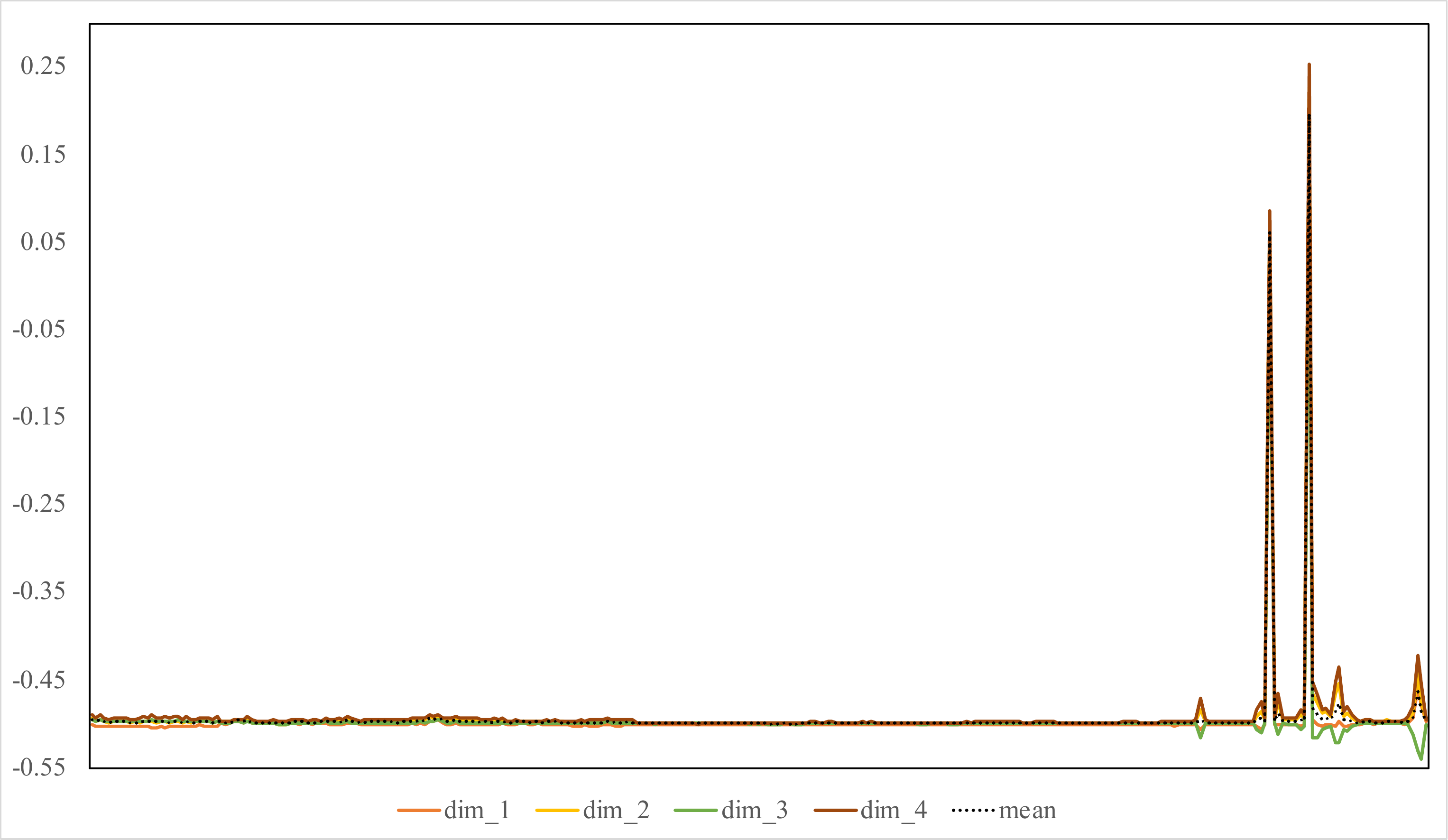}
\caption{Agreement between three primary capsules and real output capsule for fake input.}
\label{fig:agreement_fake}
\end{figure*}

\textbf{The second question} we address is about the activation of the network given real input: Given real input, does the network still focus on the same regions as fake input? A capsule networks has output capsules equal in number to the number of labels. Therefore, real inputs also trigger the capsule networks, as illustrated in Fig.~\ref{fig:viz_real}. The activation areas were similar to those for the fake inputs, mostly focused on the eyes, nose, mouth region, and facial contours. The agreement between primary capsules was the same as with the fake input. Given the behaviors in both cases, we concluded that, regardless of the input type, the primary capsules focus on specific areas and test them to determine they are original or manipulated. The results are then routed to the appropriate output capsule (real or fake one for binary classification). The algorithm then calculates the agreement for both the real and fake output capsules. The stronger the agreement for a capsule, the more certain the label.

\textbf{The third question} is about agreement between the primary capsules for time series (video) input: What is the behavior of the agreement during time? To answer this question, we tested the Capsule-Forensics network on a deepfake video and captured the activation of the real and fake capsules before the softmax function was applied. The results are plotted in Figs.~\ref{fig:agreement_real} and~\ref{fig:agreement_fake}. We took the output before the softmax function because the values are separated more and thus easier to distinguish in the plots. Each output capsule was a 4-dimensional vector, and the dynamic routing algorithm calculated agreement for all primary capsules on each dimension of the output capsules. As we can see from the two figures, agreement varied over time and differed between dimensions. Near the end of the video, there were a few frames that caused disagreements between the primary capsules. However, because we took the average of the agreements over the four dimensions, these disagreements were partially canceled out. Furthermore, the primary capsules had stronger agreement for the fake capsule than for the real capsule; therefore, the final results were still correct. It is important to note that this type of `high level fusion' is performed on latent features, unlike traditional probabilistic ensemble fusion.

\section{Evaluation}
In our evaluation of the proposed method, we first tested the improvements made to the Capsule-Forensics network: (1) using a larger input size ($300 \times 300$), (2) using dropout on training, and (3) using a larger number of primary capsules (ten). To make the results more convincing, we tested these settings on a large and challenging database -- the FaceForensics++ database~\cite{rossler2019faceforensics++}, which focuses on computer-manipulated images and videos. \textbf{\textit{After figuring out the best settings}}, we evaluated its ability to detect fully computer-generated images and presentation attacks.

For training, we used the Adam optimizer~\cite{kingma2014adam} with $\beta_1 = 0.9$, $\beta_2 = 0.999$, and a learning rate of $5\times 10^{-4}$. We used a dropout rate of 5\% and a batch size of 100 for $128 \times 128$ inputs and one of 32 for $300 \times 300$ inputs. We used the exact network architecture illustrated in Fig.~\ref{fig:capsnet}.

For comparison with other methods, depending on the database, we used three metrics.

\begin{itemize}
\item Equal error rate (EER): the value when the false acceptance rate (FAR) is equal to the false rejection rate (FRR).
\item Half total error rate: $\text{HTER} = \frac{\text{FRR} + \text{FAR}}{2}$.
\item Classification accuracy $= \frac{{TP} + \text{TN}}{\text{TP} + \text{TN} + \text{FP} + \text{FN}}$, where TP, TN, FP, and FN are true positive, true negative, false positive, and false negative, respectively.
\end{itemize}

\subsection{Detecting Computer-Manipulated Images/Videos}
\label{sec:faceforensics}
In a test to detect computer-manipulated images and videos, we used the FaceForensics++ database~\cite{rossler2019faceforensics++}, which includes three kinds of manipulations (deepfake~\cite{deepfake}, Face2Face~\cite{thies2016face2face}, and FaceSwap~\cite{rossler2019faceforensics++}) and three multiple compression levels (no compression, light compression, and heavy compression). The database was divided into a training set, a validation set, and a test set, as shown in Table~\ref{tab:faceforensics}. For the training set, we took the first 100 frames of the input video while for validation and test sets, we took only the first 10 frames. Since the Capsule-Forensics network is not limited to binary classification, we also evaluated its \textbf{multi-class classification} ability by changing the number of output capsules, from `Real' and `Fake' capsules to `Real', `Deepfakes', `Face2Face', and `FaceSwap' capsules. This modification is obvious and did not require significant changes to the network architecture.

\begin{table}[th!]
	\caption{Configuration of training, validation, and test sets based on FaceForensics++ database~\cite{rossler2019faceforensics++}.}
	\label{tab:faceforensics}
	\adjustbox{max width=\columnwidth}{%
	\centering
	\begin{tabular}{l|l|l|l}
		\multicolumn{1}{c|}{\textbf{Type}} & \multicolumn{1}{c|}{\textbf{Training set}} & \multicolumn{1}{c|}{\textbf{Validation set}} & \multicolumn{1}{c}{\textbf{Test set}} \\ \hline
		Real & \begin{tabular}[c]{@{}l@{}}$720 \times 3$ vids\\ $72,000 \times 3$ imgs\end{tabular} & \begin{tabular}[c]{@{}l@{}}$140 \times 3$ vids\\ $1,400 \times 3$ imgs\end{tabular} & \begin{tabular}[c]{@{}l@{}}$140 \times 3$ vids\\ $1,400 \times 3$ imgs\end{tabular} \\ \hline
		Deepfakes & \begin{tabular}[c]{@{}l@{}}$720 \times 3$ vids\\ $72,000 \times 3$ imgs\end{tabular} & \begin{tabular}[c]{@{}l@{}}$140 \times 3$ vids\\ $1,400 \times 3$ imgs\end{tabular} & \begin{tabular}[c]{@{}l@{}}$140 \times 3$ vids\\ $1,400 \times 3$ imgs\end{tabular} \\ \hline
		Face2Face & \begin{tabular}[c]{@{}l@{}}$720 \times 3$ vids\\ $72,000 \times 3$ imgs\end{tabular} & \begin{tabular}[c]{@{}l@{}}$140 \times 3$ vids\\ $1,400 \times 3$ imgs\end{tabular} & \begin{tabular}[c]{@{}l@{}}$140 \times 3$ vids\\ $1,400 \times 3$ imgs\end{tabular} \\ \hline
		FaceSwap & \begin{tabular}[c]{@{}l@{}}$720 \times 3$ vids\\ $72,000 \times 3$ imgs\end{tabular} & \begin{tabular}[c]{@{}l@{}}$140 \times 3$ vids\\ $1,400 \times 3$ imgs\end{tabular} & \begin{tabular}[c]{@{}l@{}}$140 \times 3$ vids\\ $1,400 \times 3$ imgs\end{tabular}
	\end{tabular}}
\end{table}

\begin{table*}[th!]
\caption{Results for all versions of Capsule-Forensics and baseline for images of various sizes except for last row showing results for videos (`light' means new version of Capsule-Forensics with only three primary capsules).}
	\label{tab:faceforensics_result}
	\adjustbox{max width=\textwidth}{%
	\centering
	\begin{tabular}{l|c|c|c|c}
		\multicolumn{1}{c|}{\textbf{Network}} & \textbf{\begin{tabular}[c]{@{}c@{}}Binary classification\\ accuracy (\%)\end{tabular}} & \textbf{\begin{tabular}[c]{@{}c@{}}Binary classification\\ EER (\%)\end{tabular}} & \textbf{\begin{tabular}[c]{@{}c@{}}Multi-class classification\\ accuracy (\%)\end{tabular}} & \textbf{\begin{tabular}[c]{@{}c@{}}Number of\\ parameters\end{tabular}} \\ \hline
		XceptionNet $(299 \times 299)$~\cite{rossler2019faceforensics++} & 91.46 & 9.98 & 91.33 & 20,811,050 \\
		Capsule-Forensics (old) $(128 \times 128)$~\cite{nguyen2019capsule} & 87.73 & 15.69 & 85.89 & 2,796,889 \\
		Capsule-Forensics (old) + Noise $(128 \times 128)$~\cite{nguyen2019capsule} & 88.11 & 15.71 & 87.12 & 2,796,889 \\ \hline
		Capsule-Forensics light $(300 \times 300)$ & 90.02 & 10.95 & 87.51 & 2,796,889 \\
		Capsule-Forensics light + Noise $(300 \times 300)$ & 91.12 & 11.60 & 87.54 & 2,796,889 \\
		Capsule-Forensics $(300 \times 300)$ & 91.65 & 11.36 & 88.51 & 3,896,638 \\
		Capsule-Forensics + Noise $(300 \times 300)$ & 91.48 & 11.62 & 89.98 & 3,896,638 \\
		Capsule-Forensics light + Dropout $(300 \times 300)$ & 91.36 & 11.61 & 89.19 & 2,796,889 \\
		Capsule-Forensics light + Dropout + Noise $(300 \times 300)$ & 91.28 & 11.38 & 88.44 & 2,796,889 \\
		Capsule-Forensics + Dropout $(300 \times 300)$ & 92.20 & 10.96 & 90.51 & 3,896,638 \\
		\textbf{Capsule-Forensics + Dropout + Noise} $\mathbf{(300 \times 300)}$ & 92.02 & 10.26 & 91.22 & 3,896,638 \\
		\hline
		\textbf{Capsule-Forensics + Dropout + Noise (video)} & 93.11 & 10.26 & 92.90 & 3,896,638 \\
	\end{tabular}}
\end{table*}

\begin{table*}[th!]
\centering
\caption{Multi-class classification accuracy for FaceForensics++ database~\cite{rossler2019faceforensics++} (\%).}
	\label{tab:result_multiclass}
	\begin{tabular}{l|c|c|c|c}
		\multicolumn{1}{c|}{\textbf{Method}} & \textbf{Real} & \textbf{Deepfakes} & \textbf{Face2Face} & \textbf{FaceSwap} \\ \hline
		XceptionNet $(299 \times 299)$~\cite{rossler2019faceforensics++} & 89.43 & 94.81 & 88.00 & 92.76 \\
		Capsule-Forensics + Dropout + Noise $(300 \times 300)$ & 89.57 & 92.17 & 90.00 & 92.79 \\
		Capsule-Forensics + Dropout + Noise (video) & 89.57 & 92.17 & 90.36 & 92.79
	\end{tabular}
\end{table*}

We used a variation of the XceptionNet~\cite{chollet2017xception} as a baseline classifier and trained it in accordance with the guidelines used by Rossler et al. ~\cite{rossler2019faceforensics++}. In this research, it achieved state-of-the-art performance on detecting manipulated images. However, the XceptionNet is a large network with more than 20 million parameters. We also compared the Capsule-Forensics with new improvements from this work (\textbf{Capsule-Forensics light} with three primary capsules and \textbf{Capsule-Forensics} with ten primary capsules) with the previous version (\textbf{Capsule-Forensics (old)})~\cite{nguyen2019capsule}, as listed in Table~\ref{tab:faceforensics_result}. The previous version included three primary capsules and had the option of adding random noise during the training process to reduce overfitting. For the new versions, one of the new settings enabled the use of a larger input size ($300 \times 300$ instead of $128 \times 128$), meaning that a larger number of important artifacts in the images. Another new setting enabled the use of more capsules (ten capsules instead of three), which increased the network capability. We call the 3-capsule version the light one. The last new setting enabled the use of dropout during the training process, which also played the role of a regularizer. This final version is called \textbf{Capsule-Forensics + Dropout + Noise} $\mathbf{(300 \times 300)}$ in Table~\ref{tab:faceforensics_result}. For simplicity, we also evaluated the final version of the Capsule-Forensics on video inputs by applying frame aggregation by calculating the average of the classification probabilities on the first ten frames (named as \textbf{Capsule-Forensics + Dropout + Noise (video)}. All versions of Capsule-Forensics were trained with 25 epochs. The results for XceptionNet and all versions of Capsule-Forensics on the test set are shown in Table~\ref{tab:faceforensics_result}.

As expected, the use of larger images improved the performance of Capsule-Forensics substantially. Regarding random noise, in our previous work~\cite{nguyen2019capsule}, most of the training sets were small, so random noise made a significant contribution. In this work, we used the first 100 frames instead of the first 10 for the training set, so the set was larger. Although the random noise did not result in improvement in all cases, it still played an important role in improving classification accuracy and reducing the EER, especially when using it was used with dropout. Increasing the number of primary capsules also helped to improve performance. Combining them together, The performance of Capsule-Forensics with both random noise and dropout was almost the same as that of XceptionNet even though the number of parameters was five time smaller, which is significant. The effects of these improvements on Capsule-Forensics performance is clearer for multi-class classification, which is more difficult than binary classification. For video input, the frame aggregation strategy increased classification accuracy for both binary and multi-class classification, as was observed in our previous work~\cite{nguyen2019capsule}.

Beyond results shown in Table~\ref{tab:faceforensics_result}, we also performed deeper analysis on multi-class classification by calculating the classification accuracy for each class. As shown in Table~\ref{tab:result_multiclass}, a Face2Face attack was the most difficult one to detect for both networks, especially for XceptionNet, with only 88.00\% accuracy. Using video inputs improved Capsule-Forensics performance even more. However, XceptionNet performed better on deepfakes detection, with accuracy of 94.81\%. For real input, both networks had slightly high false positive rates compared with their false negative rates on the manipulated inputs. Overall, Capsule-Forensics had more balanced performance for all labels than XceptionNet.

\subsection{Detecting Fully Computer-Generated Images}

In addition to testing Capsule-Forensics on computer-manipulated image, we also trained it to classify fully computer-generated images (CGIs) and photographic images (PIs). We used the dataset created by Rahmouni et al.~\cite{rahmouni2017distinguishing}. The CGI set contained 1800 high-resolution screenshots (around $1920 \times 1080$ pixels) from five photo-realistic video games. The PI set included 1800 high-resolution photographic images (around $4900 \times 3200$) selected from the RAISE dataset~\cite{dang2015raise}. We followed the prescribed protocol by training Capsule-Forensics on a $100 \times 100$ patch dataset and evaluated it on both patch and full-scale datasets. Accordingly, the input images were only $100 \times 100$. For the full-scale dataset, we used the same patch aggregation strategy by calculating the average classification probability of all patches. As shown in Table~\ref{tab:cgi_pi}, both the old and new versions of Capsule-Forensics outperformed the three other state-of-the-art classifiers and achieved 100\% classification accuracy on the large-scale dataset.

\begin{table}[th!]
	\centering
	\caption{Accuracy of state-of-the-art methods on discriminating CGIs and PIs.}	
	\label{tab:cgi_pi}
	\begin{tabular}{l|c|c}
		\multicolumn{1}{c|}{\multirow{2}{*}{\textbf{Method}}} & \multicolumn{2}{c}{\textbf{Accuracy}} \\ \cline{2-3} 
		\multicolumn{1}{c|}{} & \textbf{Patch} & \textbf{Large-scale} \\ \hline
		Rahmouni et al.~\cite{rahmouni2017distinguishing} & 89.76 & 99.30 \\
		Quan et al.~\cite{quan2018distinguishing} & 94.75 & 99.58 \\
		Nguyen et al.~\cite{nguyen2018modular} & 96.55 & 99.86 \\
		Capsule-Forensics (old)~\cite{nguyen2019capsule} & 97.00 & \textbf{100.00} \\
		Capsule-Forensics (new) & \textbf{97.05} & \textbf{100.00}
	\end{tabular}
\end{table}

\subsection{Detecting Presentation Attacks}
In addition to the experiments demonstrating the ability of Capsule-Forensics to detect images manipulated or generated by computer, we trained Capsule-Forensics on Idiap's Replay-Attack database~\cite{chingovska2012effectiveness}. Since the resolution of the videos was $320 \times 240$, we center cropped each frame to $240 \times 240$ before inputting them. As shown in Table~\ref{tab:present}, Nguyen et al.'s version~\cite{nguyen2018modular} and Capsule-Forensics achieved perfect results without any mistakes on any frames.

\begin{table}[th!]
\centering
\caption{Half total error rate (HTER) of state-of-the-art detection methods on Replay-Attack database~\cite{chingovska2012effectiveness}.}
	\label{tab:present}
	\begin{tabular}{l|c}
		\multicolumn{1}{c|}{\textbf{Method}} & \textbf{HTER (\%)} \\ \hline
		Chigovska et al.~\cite{chingovska2012effectiveness} & 17.17 \\
		de Freitas Pereira et al.~\cite{de2013can} & 08.51 \\
		Kim et al.~\cite{kim2015face} & 12.50 \\
		Yang et al.~\cite{yang2014learn} & 02.30 \\
		Menotti et al.~\cite{menotti2015deep} & 00.75 \\
		Alotabi et al.~\cite{alotaibi2017deep} & 10.00 \\
		Ito et al.~\cite{ito2017recent} & 00.43 \\
		Nguyen et al.~\cite{nguyen2018modular} & \textbf{00.00} \\
		Capsule-Forensics (old)~\cite{nguyen2019capsule} & \textbf{00.00} \\
		Capsule-Forensics (new) & \textbf{00.00}
	\end{tabular}
\end{table}

\section{Conclusion}
Our proposed Capsule-Forensics method can be applied to digital images and video forensics, including detecting computer-manipulated/generated images and videos and detecting presentation attacks. The improvements made have given Capsule-Forensics performance that is equivalent to or better than that of state-of-the-art methods on the tasks tested while using fewer parameters, which helps reduce computation cost. Detailed analysis of how the Capsule-Forensics works by visualizing the activation of each capsules and of the whole network and by analyzing the agreement between the primary capsules for video input explained the mechanism which helped the Capsule-Forensics performed well on several digital forensics tasks. With these promising results and the understanding gained from detail analysis, this work should lead to further research and development on capsule networks, not only for digital forensics but in many other areas. Future work could include application of capsule networks to time series input, not simply using frame aggregation and improvement in the generability of capsule networks, an active and challenging research topic in machine learning.

\section{Acknowledgments}
This research was supported by JSPS KAKENHI Grants JP16H06302 and JP18H04120 and by JST CREST Grant JPMJCR18A6, Japan.

\bibliographystyle{IEEEtran}
\bibliography{refs}

\begin{thebibliography}{10}
\providecommand{\url}[1]{#1}
\csname url@samestyle\endcsname
\providecommand{\newblock}{\relax}
\providecommand{\bibinfo}[2]{#2}
\providecommand{\BIBentrySTDinterwordspacing}{\spaceskip=0pt\relax}
\providecommand{\BIBentryALTinterwordstretchfactor}{4}
\providecommand{\BIBentryALTinterwordspacing}{\spaceskip=\fontdimen2\font plus
\BIBentryALTinterwordstretchfactor\fontdimen3\font minus
  \fontdimen4\font\relax}
\providecommand{\BIBforeignlanguage}[2]{{%
\expandafter\ifx\csname l@#1\endcsname\relax
\typeout{** WARNING: IEEEtran.bst: No hyphenation pattern has been}%
\typeout{** loaded for the language `#1'. Using the pattern for}%
\typeout{** the default language instead.}%
\else
\language=\csname l@#1\endcsname
\fi
#2}}
\providecommand{\BIBdecl}{\relax}
\BIBdecl

\bibitem{alexander2010digital}
O.~Alexander, M.~Rogers, W.~Lambeth, J.-Y. Chiang, W.-C. Ma, C.-C. Wang, and
  P.~Debevec, ``The digital emily project: Achieving a photorealistic digital
  actor,'' \emph{IEEE Computer Graphics and Applications}, vol.~30, no.~4, pp.
  20--31, 2010.

\bibitem{dexterstudio}
``Dexter studio,'' \url{http://dexterstudios.com/en/}, accessed: 2019-09-01.

\bibitem{karras2019style}
T.~Karras, S.~Laine, and T.~Aila, ``A style-based generator architecture for
  generative adversarial networks,'' in \emph{Conference on Computer Vision and
  Pattern Recognition (CVPR)}, 2019, pp. 4401--4410.

\bibitem{deepfake}
``Terrifying high-tech porn: Creepy 'deepfake' videos are on the rise,''
  \url{https://www.foxnews.com/tech/terrifying-high-tech-porn-creepy-deepfake-videos-are-on-the-rise},
  accessed: 2018-02-17.

\bibitem{thies2016face2face}
J.~Thies, M.~Zollhofer, M.~Stamminger, C.~Theobalt, and M.~Nie{\ss}ner,
  ``{Face2Face}: Real-time face capture and reenactment of {RGB} videos,'' in
  \emph{Conference on Computer Vision and Pattern Recognition (CVPR)}.\hskip
  1em plus 0.5em minus 0.4em\relax IEEE, 2016.

\bibitem{thies2019deferred}
J.~Thies, M.~Zollh{\"o}fer, and M.~Nie{\ss}ner, ``Deferred neural rendering:
  Image synthesis using neural textures,'' in \emph{Computer Graphics and
  Interactive Techniques (SIGGRAPH)}.\hskip 1em plus 0.5em minus 0.4em\relax
  ACM, 2019.

\bibitem{kim2018deep}
H.~Kim, P.~Garrido, A.~Tewari, W.~Xu, J.~Thies, M.~Nie{\ss}ner, P.~P{\'e}rez,
  C.~Richardt, M.~Zollh{\"o}fer, and C.~Theobalt, ``Deep video portraits,'' in
  \emph{International Conference and Exhibition on Computer Graphics and
  Interactive Techniques (SIGGRAPH)}.\hskip 1em plus 0.5em minus 0.4em\relax
  ACM, 2018.

\bibitem{averbuch2017bringing}
H.~Averbuch-Elor, D.~Cohen-Or, J.~Kopf, and M.~F. Cohen, ``Bringing portraits
  to life,'' \emph{ACM Transactions on Graphics}, 2017.

\bibitem{suwajanakorn2017synthesizing}
S.~Suwajanakorn, S.~M. Seitz, and I.~Kemelmacher-Shlizerman, ``Synthesizing
  obama: learning lip sync from audio,'' \emph{ACM Transactions on Graphics},
  2017.

\bibitem{marcel2019handbook}
S.~Marcel, M.~S. Nixon, and S.~Z. Li, \emph{Handbook of biometric
  anti-spoofing}.\hskip 1em plus 0.5em minus 0.4em\relax Springer, 2019,
  vol.~2.

\bibitem{chingovska2012effectiveness}
I.~Chingovska, A.~Anjos, and S.~Marcel, ``On the effectiveness of local binary
  patterns in face anti-spoofing,'' in \emph{International Conference of the
  Biometrics Special Interest Group (BIOSIG)}, 2012.

\bibitem{de2013can}
T.~de~Freitas~Pereira, A.~Anjos, J.~M. De~Martino, and S.~Marcel, ``Can face
  anti-spoofing countermeasures work in a real world scenario?'' in
  \emph{International Conference on Biometrics (ICB)}.\hskip 1em plus 0.5em
  minus 0.4em\relax IAPR, 2013.

\bibitem{kim2015face}
W.~Kim, S.~Suh, and J.-J. Han, ``Face liveness detection from a single image
  via diffusion speed model,'' \emph{IEEE TIP}, 2015.

\bibitem{ito2017recent}
K.~Ito, T.~Okano, and T.~Aoki, ``Recent advances in biometrics security: A case
  study of liveness detection in face recognition,'' in \emph{Asia-Pacific
  Signal and Information Processing Association Annual Summit and Conference
  (APSIPA ASC)}.\hskip 1em plus 0.5em minus 0.4em\relax IEEE, 2017.

\bibitem{george2019deep}
A.~George and S.~Marcel, ``Deep pixel-wise binary supervision for face
  presentation attack detection,'' in \emph{International Conference on
  Biometrics (ICB)}, 2019.

\bibitem{tripathy2019icface}
S.~Tripathy, J.~Kannala, and E.~Rahtu, ``Icface: Interpretable and controllable
  face reenactment using gans,'' \emph{arXiv preprint arXiv:1904.01909}, 2019.

\bibitem{yuval2019fsgan}
Y.~Nirkin, Y.~Keller, and T.~Hassner, ``Fsgan: Subject agnostic face swapping
  and reenactment,'' in \emph{International Conference on Computer Vision
  (ICCV)}.\hskip 1em plus 0.5em minus 0.4em\relax IEEE, 2019.

\bibitem{li2018ictu}
Y.~Li, M.-C. Chang, H.~Farid, and S.~Lyu, ``In ictu oculi: Exposing {AI}
  generated fake face videos by detecting eye blinking,'' \emph{arXiv preprint
  arXiv:1806.02877}, 2018.

\bibitem{korshunovvulnerability}
P.~Korshunov and S.~Marcel, ``Vulnerability assessment and detection of
  deepfake videos,'' in \emph{International Conference on Biometrics (ICB)},
  2019.

\bibitem{agarwal2019protecting}
S.~Agarwal, H.~Farid, Y.~Gu, M.~He, K.~Nagano, and H.~Li, ``Protecting world
  leaders against deep fakes,'' in \emph{Conference on Computer Vision and
  Pattern Recognition Workshops (CVPRW)}, 2019, pp. 38--45.

\bibitem{sabir2019recurrent}
E.~Sabir, J.~Cheng, A.~Jaiswal, W.~AbdAlmageed, I.~Masi, and P.~Natarajan,
  ``Recurrent convolutional strategies for face manipulation detection in
  videos,'' in \emph{Conference on Computer Vision and Pattern Recognition
  Workshops (CVPRW)}, 2019, pp. 80--87.

\bibitem{afchar2018mesonet}
D.~Afchar, V.~Nozick, J.~Yamagishi, and I.~Echizen, ``{MesoNet}: a compact
  facial video forgery detection network,'' in \emph{International Workshop on
  Information Forensics and Security (WIFS)}.\hskip 1em plus 0.5em minus
  0.4em\relax IEEE, 2018.

\bibitem{rossler2018faceforensics}
A.~R{\"o}ssler, D.~Cozzolino, L.~Verdoliva, C.~Riess, J.~Thies, and
  M.~Nie{\ss}ner, ``{FaceForensics}: A large-scale video dataset for forgery
  detection in human faces,'' \emph{arXiv preprint arXiv:1803.09179}, 2018.

\bibitem{korshunov2018speaker}
P.~Korshunov and S.~Marcel, ``Speaker inconsistency detection in tampered
  video,'' in \emph{European Signal Processing Conference (EUSIPCO)}.\hskip 1em
  plus 0.5em minus 0.4em\relax IEEE, 2018, pp. 2375--2379.

\bibitem{fridrich2012rich}
J.~Fridrich and J.~Kodovsky, ``Rich models for steganalysis of digital
  images,'' \emph{IEEE Transactions on Information Forensics and Security},
  2012.

\bibitem{rahmouni2017distinguishing}
N.~Rahmouni, V.~Nozick, J.~Yamagishi, and I.~Echizen, ``Distinguishing computer
  graphics from natural images using convolution neural networks,'' in
  \emph{International Workshop on Information Forensics and Security
  (WIFS)}.\hskip 1em plus 0.5em minus 0.4em\relax IEEE, 2017.

\bibitem{rossler2019faceforensics++}
A.~R{\"o}ssler, D.~Cozzolino, L.~Verdoliva, C.~Riess, J.~Thies, and
  M.~Nie{\ss}ner, ``Faceforensics++: Learning to detect manipulated facial
  images,'' in \emph{International Conference on Computer Vision (ICCV)}, 2019.

\bibitem{nguyen2019capsule}
H.~H. Nguyen, J.~Yamagishi, and I.~Echizen, ``Capsule-forensics: Using capsule
  networks to detect forged images and videos,'' in \emph{International
  Conference on Acoustics, Speech and Signal Processing (ICASSP)}.\hskip 1em
  plus 0.5em minus 0.4em\relax IEEE, 2019, pp. 2307--2311.

\bibitem{hinton2011transforming}
G.~E. Hinton, A.~Krizhevsky, and S.~D. Wang, ``Transforming auto-encoders,'' in
  \emph{International Conference on Artificial Neural Networks (ICANN)}.\hskip
  1em plus 0.5em minus 0.4em\relax Springer, 2011.

\bibitem{sabour2017dynamic}
S.~Sabour, N.~Frosst, and G.~E. Hinton, ``Dynamic routing between capsules,''
  in \emph{Conference on Neural Information Processing Systems (NIPS)}, 2017.

\bibitem{dale2011video}
K.~Dale, K.~Sunkavalli, M.~K. Johnson, D.~Vlasic, W.~Matusik, and H.~Pfister,
  ``Video face replacement,'' \emph{ACM Transactions on Graphics}, 2011.

\bibitem{garrido2015vdub}
P.~Garrido, L.~Valgaerts, H.~Sarmadi, I.~Steiner, K.~Varanasi, P.~Perez, and
  C.~Theobalt, ``Vdub: Modifying face video of actors for plausible visual
  alignment to a dubbed audio track,'' in \emph{Computer Graphics Forum},
  vol.~34.\hskip 1em plus 0.5em minus 0.4em\relax Wiley Online Library, 2015.

\bibitem{thies2016facevr}
J.~Thies, M.~Zollh{\"o}fer, M.~Stamminger, C.~Theobalt, and M.~Nie{\ss}ner,
  ``Facevr: Real-time facial reenactment and eye gaze control in virtual
  reality,'' \emph{ACM Transactions on Graphics}, 2018.

\bibitem{fried2019text}
O.~Fried, A.~Tewari, M.~Zollh{\"o}fer, A.~Finkelstein, E.~Shechtman, D.~B.
  Goldman, K.~Genova, Z.~Jin, C.~Theobalt, and M.~Agrawala, ``Text-based
  editing of talking-head video,'' in \emph{International Conference and
  Exhibition on Computer Graphics and Interactive Techniques (SIGGRAPH)}.\hskip
  1em plus 0.5em minus 0.4em\relax ACM, 2019.

\bibitem{zakharov2019few}
E.~Zakharov, A.~Shysheya, E.~Burkov, and V.~Lempitsky, ``Few-shot adversarial
  learning of realistic neural talking head models,'' \emph{arXiv preprint
  arXiv:1905.08233}, 2019.

\bibitem{vougioukas2019end}
K.~Vougioukas, S.~A. Center, S.~Petridis, and M.~Pantic, ``End-to-end
  speech-driven realistic facial animation with temporal gans,'' in
  \emph{Conference on Computer Vision and Pattern Recognition Workshops
  (CVPRW)}, 2019, pp. 37--40.

\bibitem{deepnude}
``New ai deepfake app creates nude images of women in seconds,''
  \url{https://www.theverge.com/2019/6/27/18760896/deepfake-nude-ai-app-women-deepnude-non-consensual-pornography},
  accessed: 2019-07-01.

\bibitem{ILSVRC15}
O.~Russakovsky, J.~Deng, H.~Su, J.~Krause, S.~Satheesh, S.~Ma, Z.~Huang,
  A.~Karpathy, A.~Khosla, M.~Bernstein, A.~C. Berg, and L.~Fei-Fei, ``{ImageNet
  Large Scale Visual Recognition Challenge},'' \emph{International Journal of
  Computer Vision}, 2015.

\bibitem{yang2014learn}
J.~Yang, Z.~Lei, and S.~Z. Li, ``Learn convolutional neural network for face
  anti-spoofing,'' \emph{arXiv preprint arXiv:1408.5601}, 2014.

\bibitem{menotti2015deep}
D.~Menotti, G.~Chiachia, A.~Pinto, W.~R. Schwartz, H.~Pedrini, A.~X. Falcao,
  and A.~Rocha, ``Deep representations for iris, face, and fingerprint spoofing
  detection,'' \emph{IEEE Transactions on Information Forensics and Security},
  2015.

\bibitem{raghavendra2017transferable}
R.~Raghavendra, K.~B. Raja, S.~Venkatesh, and C.~Busch, ``Transferable
  deep-{CNN} features for detecting digital and print-scanned morphed face
  images,'' in \emph{Conference on Computer Vision and Pattern Recognition
  Workshop (CVPRW)}.\hskip 1em plus 0.5em minus 0.4em\relax IEEE, 2017.

\bibitem{mehtacrafting}
S.~Mehta, A.~Uberoi, A.~Agarwal, M.~Vatsa, and R.~Singh, ``Crafting a panoptic
  face presentation attack detector,'' in \emph{International Conference on
  Biometrics (ICB)}, 2019.

\bibitem{muhammad2019face}
U.~Muhammad and A.~Hadid, ``Face anti-spoofing using hybrid residual learning
  framework,'' in \emph{International Conference on Biometrics (ICB)}, 2019.

\bibitem{alotaibi2017deep}
A.~Alotaibi and A.~Mahmood, ``Deep face liveness detection based on nonlinear
  diffusion using convolution neural network,'' \emph{Signal, Image and Video
  Processing}, 2017.

\bibitem{costa2019generalized}
A.~Costa-Pazo, D.~Jim{\'e}nez-Cabello, E.~Vazquez-Fernandez, J.~L. Alba-Castro,
  and R.~J. L{\'o}pez-Sastre, ``Generalized presentation attack detection: a
  face anti-spoofing evaluation proposal,'' in \emph{International Conference
  on Biometrics (ICB)}, 2019.

\bibitem{jaiswal2019ropad}
A.~Jaiswal, S.~Xia, I.~Masi, and W.~AbdAlmageed, ``Ropad: Robust presentation
  attack detection through unsupervised adversarial invariance,'' in
  \emph{International Conference on Biometrics (ICB)}, 2019.

\bibitem{nikisins2019domain}
O.~Nikisins, A.~George, and S.~Marcel, ``Domain adaptation in multi-channel
  autoencoder based features for robust face anti-spoofing,'' in
  \emph{International Conference on Biometrics (ICB)}, 2019.

\bibitem{wangimproving}
G.~Wang, H.~Han, S.~Shan, and X.~Chen, ``Improving cross-database face
  presentation attack detection via adversarial domain adaptation,'' in
  \emph{International Conference on Biometrics (ICB)}, 2019.

\bibitem{cozzolino2018forensictransfer}
D.~Cozzolino, J.~Thies, A.~R{\"o}ssler, C.~Riess, M.~Nie{\ss}ner, and
  L.~Verdoliva, ``Forensictransfer: Weakly-supervised domain adaptation for
  forgery detection,'' \emph{arXiv preprint arXiv:1812.02510}, 2018.

\bibitem{nguyen2019multi}
H.~H. Nguyen, F.~Fang, J.~Yamagishi, and I.~Echizen, ``Multi-task learning for
  detecting and segmenting manipulated facial images and videos,'' in
  \emph{International Conference on Biometrics: Theory,Applications and Systems
  (BTAS)}.\hskip 1em plus 0.5em minus 0.4em\relax IEEE, 2019.

\bibitem{fatemifar2019combining}
S.~Fatemifar, M.~Awais, S.~Rahimzadeh~Arashloo, and J.~Kittler, ``Combining
  multiple one-class classifiers for anomaly based face spoofing attack
  detection,'' in \emph{International Conference on Biometrics (ICB)}, 2019.

\bibitem{cozzolino2017recasting}
D.~Cozzolino, G.~Poggi, and L.~Verdoliva, ``Recasting residual-based local
  descriptors as convolutional neural networks: an application to image forgery
  detection,'' in \emph{Workshop on Information Hiding and Multimedia Security
  (IH\&MMSEC)}.\hskip 1em plus 0.5em minus 0.4em\relax ACM, 2017.

\bibitem{bayar2016deep}
B.~Bayar and M.~C. Stamm, ``A deep learning approach to universal image
  manipulation detection using a new convolutional layer,'' in \emph{Workshop
  on Information Hiding and Multimedia Security (IH\&MMSEC)}.\hskip 1em plus
  0.5em minus 0.4em\relax ACM, 2016.

\bibitem{quan2018distinguishing}
W.~Quan, K.~Wang, D.-M. Yan, and X.~Zhang, ``Distinguishing between natural and
  computer-generated images using convolutional neural networks,'' \emph{IEEE
  Transactions on Information Forensics and Security}, 2018.

\bibitem{zhou2017two}
P.~Zhou, X.~Han, V.~I. Morariu, and L.~S. Davis, ``Two-stream neural networks
  for tampered face detection,'' in \emph{Conference on Computer Vision and
  Pattern Recognition Workshop (CVPRW)}.\hskip 1em plus 0.5em minus 0.4em\relax
  IEEE, 2017.

\bibitem{nguyen2018modular}
H.~H. Nguyen, N.-D.~T. Tieu, H.-Q. Nguyen-Son, V.~Nozick, J.~Yamagishi, and
  I.~Echizen, ``Modular convolutional neural network for discriminating between
  computer-generated images and photographic images,'' in \emph{International
  Conference on Availability, Reliability and Security (ARES)}.\hskip 1em plus
  0.5em minus 0.4em\relax ACM, 2018.

\bibitem{wang2019detecting}
S.-Y. Wang, O.~Wang, A.~Owens, R.~Zhang, and A.~A. Efros, ``Detecting
  photoshopped faces by scripting photoshop,'' in \emph{International
  Conference on Computer Vision (ICCV)}.\hskip 1em plus 0.5em minus 0.4em\relax
  IEEE, 2019.

\bibitem{marra2018detection}
F.~Marra, D.~Gragnaniello, D.~Cozzolino, and L.~Verdoliva, ``Detection of
  gan-generated fake images over social networks,'' in \emph{Conference on
  Multimedia Information Processing and Retrieval (MIPR)}.\hskip 1em plus 0.5em
  minus 0.4em\relax IEEE, 2018, pp. 384--389.

\bibitem{marra2019gans}
F.~Marra, D.~Gragnaniello, L.~Verdoliva, and G.~Poggi, ``Do gans leave
  artificial fingerprints?'' in \emph{Conference on Multimedia Information
  Processing and Retrieval (MIPR)}.\hskip 1em plus 0.5em minus 0.4em\relax
  IEEE, 2019, pp. 506--511.

\bibitem{bappy2017exploiting}
J.~H. Bappy, A.~K. Roy-Chowdhury, J.~Bunk, L.~Nataraj, and B.~Manjunath,
  ``Exploiting spatial structure for localizing manipulated image regions,'' in
  \emph{International Conference on Computer Vision (ICCV)}, 2017, pp.
  4970--4979.

\bibitem{zhou2018learning}
P.~Zhou, X.~Han, V.~I. Morariu, and L.~S. Davis, ``Learning rich features for
  image manipulation detection,'' in \emph{Conference on Computer Vision and
  Pattern Recognition (CVPR)}, 2018, pp. 1053--1061.

\bibitem{bappy2019hybrid}
J.~H. Bappy, C.~Simons, L.~Nataraj, B.~Manjunath, and A.~K. Roy-Chowdhury,
  ``Hybrid lstm and encoder-decoder architecture for detection of image
  forgeries,'' \emph{IEEE Transactions on Image Processing}, 2019.

\bibitem{long2015fully}
J.~Long, E.~Shelhamer, and T.~Darrell, ``Fully convolutional networks for
  semantic segmentation,'' in \emph{Conference on Computer Vision and Pattern
  Recognition (CVPR)}, 2015, pp. 3431--3440.

\bibitem{badrinarayanan2017segnet}
V.~Badrinarayanan, A.~Kendall, and R.~Cipolla, ``Segnet: A deep convolutional
  encoder-decoder architecture for image segmentation,'' \emph{IEEE
  Transactions on Pattern Analysis and Machine Intelligence}, vol.~39, no.~12,
  pp. 2481--2495, 2017.

\bibitem{hinton2018matrix}
G.~E. Hinton, S.~Sabour, and N.~Frosst, ``Matrix capsules with {EM} routing,''
  in \emph{International Conference on Learning Representations Workshop
  (ICLRW)}, 2018.

\bibitem{xi2017capsule}
E.~Xi, S.~Bing, and Y.~Jin, ``Capsule network performance on complex data,''
  \emph{arXiv preprint arXiv:1712.03480}, 2017.

\bibitem{xiang2018ms}
C.~Xiang, L.~Zhang, Y.~Tang, W.~Zou, and C.~Xu, ``Ms-capsnet: A novel
  multi-scale capsule network,'' \emph{IEEE Signal Processing Letters},
  vol.~25, no.~12, pp. 1850--1854, 2018.

\bibitem{bahadori2018spectral}
M.~T. Bahadori, ``Spectral capsule networks,'' in \emph{International
  Conference on Learning Representations (ICLR)}, 2018.

\bibitem{iesmantas2018convolutional}
T.~Iesmantas and R.~Alzbutas, ``Convolutional capsule network for
  classification of breast cancer histology images,'' in \emph{International
  Conference Image Analysis and Recognition ( ICIAR)}.\hskip 1em plus 0.5em
  minus 0.4em\relax Springer, 2018, pp. 853--860.

\bibitem{jaiswal2018capsulegan}
A.~Jaiswal, W.~AbdAlmageed, Y.~Wu, and P.~Natarajan, ``Capsulegan: Generative
  adversarial capsule network,'' in \emph{European Conference on Computer
  Vision (ECCV)}, 2018, pp. 0--0.

\bibitem{yang2018investigating}
M.~Yang, W.~Zhao, J.~Ye, Z.~Lei, Z.~Zhao, and S.~Zhang, ``Investigating capsule
  networks with dynamic routing for text classification,'' in \emph{Conference
  on Empirical Methods in Natural Language Processing (EMNLP)}, 2018, pp.
  3110--3119.

\bibitem{simonyan2014very}
K.~Simonyan and A.~Zisserman, ``Very deep convolutional networks for
  large-scale image recognition,'' in \emph{International Conference on
  Learning Representations (ICLR)}, 2015.

\bibitem{kingma2014adam}
D.~P. Kingma and J.~Ba, ``Adam: A method for stochastic optimization,'' in
  \emph{International Conference on Learning Representations (ICLR)}, 2015.

\bibitem{uozbulak_pytorch_vis_2019}
U.~Ozbulak, ``Pytorch cnn visualizations,''
  \url{https://github.com/utkuozbulak/pytorch-cnn-visualizations}, 2019.

\bibitem{selvaraju2017grad}
R.~R. Selvaraju, M.~Cogswell, A.~Das, R.~Vedantam, D.~Parikh, and D.~Batra,
  ``Grad-cam: Visual explanations from deep networks via gradient-based
  localization,'' in \emph{International Conference on Computer Vision
  (ICCV)}.\hskip 1em plus 0.5em minus 0.4em\relax IEEE, 2017, pp. 618--626.

\bibitem{chollet2017xception}
F.~Chollet, ``Xception: Deep learning with depthwise separable convolutions,''
  in \emph{Conference on Computer Vision and Pattern Recognition (CVPR)}.\hskip
  1em plus 0.5em minus 0.4em\relax IEEE, 2017.

\bibitem{dang2015raise}
D.-T. Dang-Nguyen, C.~Pasquini, V.~Conotter, and G.~Boato, ``Raise: A raw
  images dataset for digital image forensics,'' in \emph{Multimedia Systems
  Conference (MMSys)}.\hskip 1em plus 0.5em minus 0.4em\relax ACM, 2015, pp.
  219--224.

\end{thebibliography}

\end{document}